# Multi Jet Fusion of Nylon-12: A Viable Method to 3D-print Concentric Tube Robots?

K. Picho, B. Persons, J.F. d'Almeida, N.E. Pacheco, C. Reynolds, L. Fichera

*Department of Robotics Engineering, Worcester Polytechnic Institute, USA*
kmpicho@wpi.edu

## INTRODUCTION

Concentric Tube Robots (CTRs) are needle-sized flexible manipulators well suited for minimally invasive surgery. The body of a CTR consists of a set of pre-curved tubes nested telescopically within each other, as shown in Fig. 1. Each tube can be translated and rotated independently. During actuation, the tubes interact elastically with each other, creating tentacle-like bending motions. CTRs are usually made of superelastic Nickel-Titanium (Nitinol), a metal alloy capable of withstanding large amounts of strain (typically up to 6-8%) without undergoing plastic deformation. Strain recovery is key for the operation of a CTR, as it enables the body of the robot to undergo substantial yet reversible bending.

In recent years, different groups have investigated the viability of 3D-printing CTRs using materials other than Nitinol [1-2]. This research is motivated by the fact that Nitinol can be a challenging material to work with: to build CTRs, Nitinol tubing must first be formed into a prescribed curved shape through an annealing treatment, which, as previous work has shown, can be complex, time-consuming, and error-prone [3]. Correct execution of this process requires specialized expertise as well as equipment that may not be readily available in a robotics laboratory. 3D-printing promises to overcome these challenges by providing a way to rapidly prototype CTRs. This would not only benefit CTR research, but it would also enable new appealing capabilities, such as the ability to create patient- or procedure-specific robots.

In this paper, we present a study on the viability of fabricating CTRs using Multi Jet Fusion (MJF) of Nylon-12, a type of elastic polymer commonly used in additive manufacturing [4]. We note that Nylon-12 was already evaluated for the purpose of building CTRs in [2], but fabrication was performed with Selective Laser Sintering (SLS), which produced unsatisfactory results. Our study is the first study to evaluate the suitability of MJF to 3D-print CTRs.

## MATERIALS AND METHODS

MJF is a novel additive manufacturing technology developed by Hewlett-Packard (Palo Alto, CA, USA) [4]. Parts are built by depositing successive layers of material powder, which are then heated and fused. This process is similar to SLS, but powder heating uses infrared light (as opposed to a laser in SLS), in combination with chemical agents that facilitate heat absorption. MJF can create parts with mechanical properties and finish similar to those printed with SLS [4]. Unlike SLS, however, the unique treating agents that MJF uses help to create tiny features. Thus, MJF is capable of higher dimensional accuracy, finer resolution, and smaller wall thicknesses, which all significantly aid in the printing of CTRs. For this study, MJF fabrication of the tubes was outsourced to Protolabs (Maple Plain, MN, USA).

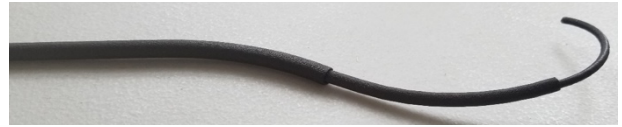

**Fig. 1:** 3D-printed Concentric Tube Robot.

**Stress-Strain Characterization**: As a first step in our study, we performed a tensile test to experimentally characterize the stress-strain curve of Nylon-12 parts created with MJF. This test was carried out according to the ASTM D638 protocol, using specially designed "dog-bone" shaped parts and an Instron 5500R Universal Testing Machine (Instron, Norwood, MA, USA). This test enabled us to estimate the linear elastic range of the material and the Young's Modulus, which are necessary to model the mechanics of CTRs [5].

**Fatigue Testing**: Prior work found Nylon-12 tubes printed via SLS to be brittle and prone to breaking [2]. To verify if Nylon-12 tubes created with MJF can withstand multiple bending cycles, we performed a fatigue test. This test used a tube with an outer diameter of 3.2 mm, a wall thickness of 0.6 mm, and a 28.26 mm radius of curvature. The tube was straightened by pulling it inside a straight hollow shaft and then pushed back out. This process was automated by attaching the proximal end of the tube to a motorized stage and repeated for 200 cycles. Photographs were taken every 10 cycles to document the conditions of the tube.

**In-Plane Bending Verification**: We performed an experiment to verify if the in-plane bending model for concentric tubes proposed by Webster *et al.* in [5] can be used with the tubes considered in this study. Briefly, the model in [5] assumes that concentric tubes apply moments to one another by virtue of their different pre-curvatures and bending stiffnesses. If two concentric tubes are aligned so that they bend in the same plane, their equilibrium curvature is simply described by $\kappa = \frac{E_1 I_1 k_1 + E_2 I_2 k_2}{E_1 I_1 + E_2 I_2}$, where the subscripts indicate the tube number, $E$ is the Young's Modulus, $k$ is the pre-curvature, and $I$ is the second moment of area.

The experiment is illustrated in Fig. 2: two tubes are nested inside each other so that their curved sections overlap completely, and the equilibrium curvature is then estimated from images. The experiment used the combinations of inner/outer tubes listed in Table 1. All the tubes used in these trials have a wall thickness of 0.6 mm and a distal curved section of 50 mm. Each trial was repeated 11 times.

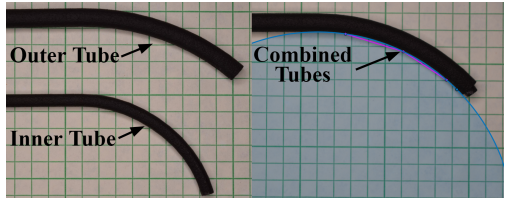

**Fig. 2:** In-plane bending test. Photographs of the tubes are taken before and after insertion with a Nikon D5100 DSLR camera. The equilibrium curvature is measured by manually digitizing three points along the curve, and then using a circle fit in MATLAB to estimate the corresponding radius.

## RESULTS

**Stress-Strain Characterization**: Fig. 3 shows the stress-strain curve estimated in the tensile test. The curve increases linearly until approximately 2% strain before entering the region of plastic deformation. Breakage of the dog bone part occurred at 17.7% strain. The Young's Modulus was estimated to be 1.51 GPa. These results are consistent with those reported in other studies [4].

**Fatigue Testing**: The tube was able to withstand all 200 bending cyles without any macroscopic or otherwise apparent sign of breakage. We did observe a progressive loss of pre-curvature of the tube, as documented in Fig. 4. We attribute the loss of curvature to plastic deformation: as can be verified using Eq. (1) in [5], the strain experienced by the material in the tube during straightening reached nearly 6%.

**In-Plane Bending Experiment**: Results are reported in Table 1. The in-plane bending model from [5] was able to predict the radius of curvature of the combined tubes, with RMSEs ranging between 0.3 and 1.9 mm.

## DISCUSSION

Results suggest that MJF of Nylon-12 may be a viable process to build CTRs. Tubes can be printed in small diameters (the smallest tube in this paper has outer diameter equal to 2.2 mm) and can withstand hundreds of bending cycles without breaking. Although the maximum recoverable strain was not determined, it would be reasonable to assume a value between 2-3% based on the stress-strain curve in Fig. 3. This value is lower than what is achievable with Nitinol (6-8%), impliying that CTRs made of Nylon-12 will not be able to achieve the same range of bending motions as those made of Nitinol. This was expected, as superelastic Nitinol is and remains the most suitable material for CTRs. MJF of Nylon-12 offers the convenience to rapidly prototype CTRs at the expense of limiting the robot's reachable workspace. One important limitation of this study is that it did not investigate the mechanics of concentric Nylon-12 tubes subject to rotation. We plan to conduct a complete verification of the CTR kinematics model in [5] in future work.

**Table 1:** Experimental conditions and results for the in-plane bending experiments. OD: Outer Diameter. R: Radius of Curvature. R*: Equilibrium Radius of Curvature predicted by the model in [5]. RMSE: Root-Mean-Square Error. All units are in millimeters.

| Trial | Outer Tube | | Inner Tube | | Combined Tubes | |
|---|---|---|---|---|---|---|
| | OD | R | OD | R | R* | RMSE |
| 1 | 3.8 | 69.0 | 2.2 | 21.9 | 54.7 | 1.7 |
| 2 | 3.8 | 39.9 | 2.2 | 22.1 | 36.4 | 0.8 |
| 3 | 5.4 | 68.6 | 3.8 | 43.2 | 64.0 | 1.9 |
| 4 | 5.4 | 34.5 | 3.8 | 70.9 | 36.8 | 0.3 |

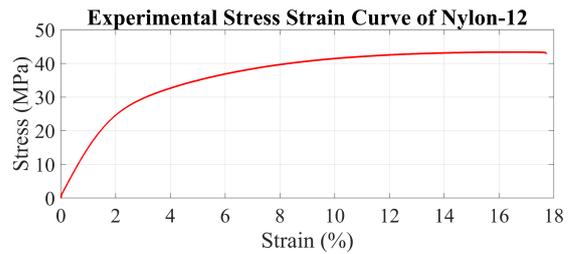

**Fig. 3:** Stress-strain curve for MJF-printed Nylon-12.

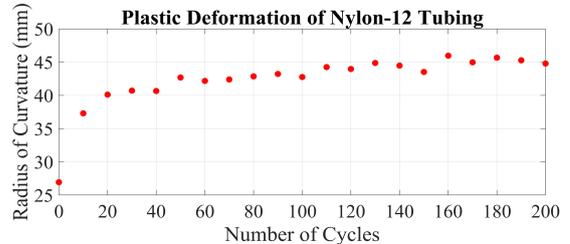

**Fig. 4:** Progressive loss of pre-curvature observed during fatigue testing.

The dimensions and pre-curvatures of the component tubes in a CTR are critical design parameters, as they determine the robot's reachable workspace. Several groups have recently explored algorithms that produce optimal designs for a given application. Until now, however, these algorithms have been of limited application due to the difficulty of sourcing Nitinol tubing with specific prescribed dimensions and pre-curvatures. By enabling the creation of tubes in virtually any arbitrary dimensions and curvatures, MJF promises to overcome these challenges and open up the way to the creation of procedure- and patient-specific robots.

## REFERENCES

[1] Amanov E, Nguyen TD, Burgner-Kahrs J. Additive manufacturing of patient-specific tubular continuum manipulators. SPIE Medical Imaging 2015.

[2] Morimoto TK, Okamura AM. Design of 3-D printed concentric tube robots. IEEE Transactions on Robotics. 2016 Sep 23;32(6):1419-30.

[3] Gilbert HB, Webster RJ. Rapid, reliable shape setting of superelastic nitinol for prototyping robots. IEEE robotics and automation letters. 2015 Dec 11;1(1):98-105.

[4] Cai C, et al. Comparative study on 3D printing of polyamide 12 by selective laser sintering and multi jet fusion. Journal of Materials Processing Technology. 2021 Feb 1;288:116882.

[5] Webster RJ, Romano JM, Cowan NJ. Mechanics of precurved-tube continuum robots. IEEE Transactions on Robotics. 2008 Nov 11;25(1):67-78.